\documentclass[a4paper]{article}

\usepackage{INTERSPEECH_v2}

\usepackage{tikz}
\usetikzlibrary{bayesnet}
\renewcommand{\edge}[3][]{ %
  \foreach \x in {#2} { %
    \foreach \y in {#3} { %
      \path (\x) edge [->,#1] (\y) ;%
    } ;
  } ;
}

\usetikzlibrary{arrows,shapes,backgrounds,positioning,fit}

\usepackage{pgfplots}
\pgfplotsset{compat=1.12}

\title{A Generative Model for Score Normalization in Speaker Recognition}
\name{Albert Swart and Niko Br\"ummer}
\address{Nuance Communications, Inc. (South Africa)}
\email{albert.swart@nuance.com, niko.brummer@nuance.com}

\def\R{\mathbb{R}}
\DeclareMathOperator{\logit}{logit}

\def\Sset{\mathcal{S}}
\def\Eset{\mathcal{E}}
\def\Tset{\mathcal{T}}

\def\Xset{\mathcal{X}}
\def\Yset{\mathcal{Y}}

\def\xvec{\mathbf{x}}
\def\yvec{\mathbf{y}}
\def\zvec{\mathbf{z}}

\begin{document}

\tikzstyle{every picture}+=[remember picture]

\tikzstyle{cbox} = [rectangle,draw=blue!100,thick,align=center,rounded corners = 3pt]
\tikzstyle{lbox} = [rectangle,draw=blue!100,thick,align=left,rounded corners = 3pt]
\tikzstyle{ccircle} = [circle,draw=blue!100,thick,align=center,inner sep = 0]
\tikzstyle{ctext} = [rectangle,align=center,inner sep = 4pt]
\tikzstyle{ltext} = [rectangle,align=left,inner sep = 4pt]
\tikzstyle{solder} = [circle,draw,fill,inner sep = 0, minimum size = 3pt]

\maketitle
\begin{abstract}
We propose a theoretical framework for thinking about score normalization, which confirms that normalization is not needed under (admittedly fragile) ideal conditions. If, however, these conditions are not met, e.g.\ under data-set shift between training and runtime, our theory reveals dependencies between scores that could be exploited by strategies such as score normalization. Indeed, it has been demonstrated over and over experimentally, that various ad-hoc score normalization recipes do work. We present a first attempt at using probability theory to design a generative score-space normalization model which gives similar improvements to ZT-norm on the text-dependent RSR 2015 database.
\end{abstract}

\section{Introduction}
Speaker recognition researchers have a love-hate relationship with score normalization. Purists regard it as a kludge.\footnote{A kludge is a workaround or quick-and-dirty solution that is clumsy, inelegant, inefficient, difficult to extend and hard to maintain.} Many researchers believe efforts are better spent elsewhere.\footnote{Patrick Kenny, in a personal communication: ``The score domain is an impoverished domain.''} However, score normalization is often the only thing that helps in practice to reduce error-rates under non-ideal conditions, such as the data-set shift encountered in the most recent NIST SRE'16 evaluation~\cite{NPT_SRE16,ABC_SRE16}.

Various flavours of score normalization have been published, for example T-norm~\cite{t-norm}, adaptive T-norm~\cite{at-norm}, ZT-norm~\cite{zt-norm}, S-norm~\cite{ht-plda} and adaptive S-norm~\cite{as-norm}. The sharp intuition and clever engineering that have contributed to these recipes should not be undervalued---after all, they work and get the job done. However, we feel there is still a lack of theory for designing score normalization solutions. Indeed, it is often regarded as an art rather than a science---see for example ``The awe and mystery of T-norm''~\cite{awe_and_mystery}. The only informal theory seems to be based on the observation that the distributions of scores originating from different test segments (or different enrollments) can sometimes be mutually misaligned. Score normalization seeks to re-align these distributions via scaling and shifting.

In this paper, we make the following contributions. We explain why score normalization is not needed under ideal circumstances, when the recognition model is a good match for the data at runtime. We then motivate why score normalization could help under more challenging conditions, when there is a mismatch between model and data. Finally, we venture into a first attempt at using probability theory to design a new score normalization recipe. We do this by defining a simple generative score-space model that uses hidden variables to induce dependency between the trial-at-hand and some cohort scores. Once the model has been defined, probability theory does the rest of the job to find the normalization recipe. We conclude with experiments on RSR 2015 to demonstrate that our approach has practical merit.

\section{Traditional score normalization}
We shall confine ourselves to the canonical speaker recognition problem, where a \emph{trial}, $(e,t)$, is scored via a function $s(e,t)\to\R$, where $e$ represents the enrollment speech of some speaker of interest, and $t$ represents the test-speech, which may or may not be of that speaker. The speaker recognition system processes enrollment and test speech to extract the representations $e$ and $t$. It also implements the scoring function $s(e,t)$. In some systems, such as i-vector PLDA~\cite{ht-plda,garcia2011lengthnorm}, $e$ and $t$ have the same form (often just i-vectors) and the scoring function is symmetric: $s(e,t)=s(t,e)$. In other systems, the representations for $e$ and $t$ differ, which requires an asymmetric scoring function.

Symmetric scoring functions can be normalized with symmetric normalization recipes such as S-norm---while asymmetric normalization recipes, such as T-norm and ZT-norm can be applied to both symmetric and asymmetric scoring functions. In the interest of generality below, we assume asymmetric scoring. For later reference, we briefly summarize two well-known score normalization recipes.

\subsection{T-norm}
We denote the \emph{trial-at-hand} as $(\tilde e,\tilde t)$ and the corresponding \emph{raw score}, as $\tilde s = s(\tilde e,\tilde t)$. The relatively simple T-norm~\cite{t-norm} makes use of a \emph{cohort}, $\Eset = \{e'_i\}_{i=1}^N$, populated by enrollments, $e'_i$, of $N$ other speakers, which we assume to be all different from the speaker(s) present in the trial-at-hand. The \emph{cohort scores}:
\begin{align}
\Sset_{\tilde t}&=\bigl\{s(e'_i,\tilde t)\bigr\}_{i=1}^N
\end{align}
are formed by scoring $\tilde t$ against the cohort. The \emph{normalized score} is formed as:
\begin{align}
\label{eq:tnorm}
s^*(\tilde e, \tilde t) &= \frac{s(\tilde e, \tilde t)-\mu(\tilde t)}{\sigma(\tilde t)}
\end{align}
where $\mu(\tilde t)$ and $\sigma(\tilde t)$ are respectively the mean and standard deviation of $\Sset_{\tilde t}$.

\subsection{ZT-norm}
\label{sec:ztnorm}
ZT-norm is more complex. It scores the trial-at-hand against two different cohorts and also scores the cohorts against each other. One cohort is populated by enrollments: $\Eset = \{e'_i\}_{i=1}^N$, and the other is populated by test representations: $\Tset = \{t'_i\}_{i=1}^M$. Three sets of cohort scores are required: $\tilde e$ scored against $\Tset$ and $\tilde t$ against $\Eset$:
\begin{align*}
\Sset_{\tilde e}&=\bigl\{s(\tilde e,t'_i)\bigr\}_{i=1}^M, &
\Sset_{\tilde t}&=\bigl\{s(e'_j,\tilde t)\bigr\}_{j=1}^N,
\intertext{and then also $\Tset$ against $\Eset$ for a matrix of \emph{inter cohort scores}:}
\Sset_I &= \bigl\{s(e'_i,t'_j)\bigr\}_{i,j=1}^{M,N}
\end{align*}
The zt-normalized score is computed by sequentially composing two normalization steps of the form~\eqref{eq:tnorm}. See~\cite{zt-norm,shum2010cosinetid} for details. Of interest here, is that the normalized score is a function of $\tilde s$, as well as all the cohort scores, $\Sset_{\tilde e},\Sset_{\tilde t},\Sset_I$. Adaptive score normalizations use similar score sets as input~\cite{at-norm,as-norm}.

\section{Theory and motivation}
We present a theoretical motivation for doing score normalization, based on conditional independence analysis. The basic question is: \emph{Given the raw score, are the cohort scores also relevant to better infer the speaker hypothesis for the trial-at-hand?}

For generality, let us consider the full set of cohort scores, $\Sset_{\tilde t}$, $\Sset_{\tilde e}$ and $\Sset_I$ as used by ZT-norm and adaptive score normalization.

\begin{figure}[htb!]
  \centering
	\begin{tikzpicture}
	\node[const, inner sep = 3pt] (Theta) {$\Theta$};
	\node[obs, right = of Theta] (tildes) {$\tilde s$};
	\node[latent, above = of tildes] (tildee) {$\tilde e$};
	\node[latent, below = of tildes] (tildet) {$\tilde t$};
	\node[obs, left = of Theta] (X) {$\Sset_I$};
	\node[latent, above = of X] (T) {$\Tset$};
	\node[latent, below = of X] (E) {$\Eset$};
	\node[obs] (St) at(tildet -| Theta) {$\Sset_{\tilde t}$};
	\node[obs] (Se) at(tildee -| Theta) {$\Sset_{\tilde e}$};
	\node[latent, right = of tildet] (tildeh) {$\tilde h$};
	\node[const, inner sep = 5pt, above = of tildeh] (pi)  {$\pi$};
	\edge{pi}{tildeh};
	\edge{tildeh}{tildet};
	\edge{tildee,tildet}{tildes};
	\edge{Theta}{tildee,tildet};
	\edge{E,T}{X};
	\edge{Theta}{E,T};
	\edge{T,tildee}{Se};
	\edge{E,tildet}{St};
	\edge[bend left = 60]{tildee}{tildet};
	\end{tikzpicture}
  \caption{Graphical model analysis of score normalization.}
  \label{fig:gm1}
\end{figure}
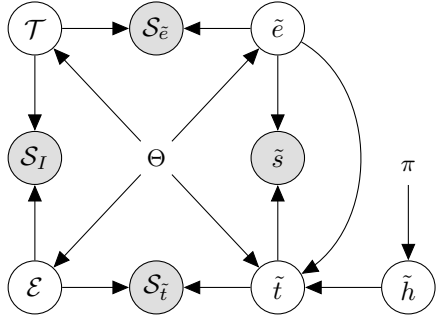

In figure~\ref{fig:gm1}, we use graphical model notation~\cite{PRML} to reason about conditional independence relationships. The shaded nodes represent the various observed scores, the cohort scores, as well as the score for the trial-at-hand, $\tilde s$. These scores will be our inputs for computing the normalized score. Hypothesis labels for all cohort scores are assumed given (traditionally all non-targets).

At normalization time, the original data for the trial-at-hand $\tilde e, \tilde t$ and the cohorts, $\Eset,\Tset$ are no longer available and are now hidden,\footnote{If we consider the data as given, this blocks dependency between $\tilde h$ and the cohort scores and score normalization no longer applies.} as indicated by clear circles. The object of the whole exercise is to infer the value of the \emph{hidden speaker hypothesis} for the trial-at-hand:\footnote{Under $\texttt{target}$, $\tilde e$ and $\tilde t$ come from the same speaker. Under $\texttt{non-target}$, they come from different sepakers.}
$$\tilde h\in\{\texttt{target},\texttt{non-target}\}$$
Let $\Theta$ denote the parameters of a probabilistic generative model that is assumed to have generated all of the data. For now, we consider $\Theta$ as given. A hypothesis prior, $$\pi = P(\tilde h =\texttt{target}\mid \pi)$$ is also given. Let us now apply the rules for dependency in graphical models~\cite{PRML} to see if we can find dependency between $\tilde h$ and the cohort scores. That is, we want to know if the computation $P(\tilde h \mid \tilde s,\pi)$ suffices for inference of $\tilde h$, or if we need \emph{everything} that is given: $P(\tilde h \mid \tilde s,\Sset_{\tilde t},\Sset_{\tilde e}, \Sset_I, \pi,\Theta)$.

Dependency can `flow' along or against any arrow, but can be `blocked' at some nodes, depending on whether the nodes are observed or not. Dependency at any observed node, $O$, is blocked along paths of the forms $\to O \to$ and $\gets O \to$. Dependency is also blocked along paths of the form $\to N \gets$ for any node $N$, if neither $N$, nor any of its descendants are observed. When these rules are applied to figure~\ref{fig:gm1}, we find that: \emph{Yes, there are dependency paths from $\tilde h$ to $\Theta$ and to all the cohort scores.} At first glance it therefore looks as if we should always be doing score normalization. But this ignores the special nature of the scores that are computed under ideal circumstances.

\subsection{Ideal circumstances}
Perusal of figure~\ref{fig:gm1} shows that as long as $\Theta$ is given, we have:
\begin{align}
\label{eq:indep}
P(\tilde h\mid \tilde e, \tilde t, \tilde s, \Sset_{\tilde e}, \Sset_{\tilde t},\Sset_I,\pi,\Theta)
&= P(\tilde h\mid \tilde e, \tilde t, \pi,\Theta)
\end{align}
Intuitively, cohorts $\Eset$ and $\Tset$ (and any scores computed from them) are independent observations of speakers different from the speaker(s) in the trial-at-hand. If $\Theta$ is given, data from the cohorts are not needed to learn more about $\Theta$, and the cohorts will be of no further help in the trial-at-hand. Now consider the \emph{ideal} score, namely the \emph{log-likelihood-ratio}:
\begin{align}
\label{eq:llr}
\tilde s &= s(\tilde e,\tilde t) = \log\frac{P(\tilde e,\tilde t\mid\tilde h=\texttt{target},\Theta)}{P(\tilde e,\tilde t\mid\tilde h=\texttt{non-target},\Theta)}
\end{align}
in which case we can apply Bayes' rule to see:
\begin{align}
\label{eq:suff}
\begin{split}
&P(\tilde h=\texttt{target}\mid\tilde e, \tilde t, \pi,\Theta) \\
&= \sigma\bigl(\tilde s+\logit\pi\bigr) \\
&= P(\tilde h=\texttt{target}\mid\tilde s,\pi)
\end{split}
\end{align}
where $\sigma(s)=\frac{1}{1+e^{-s}}$ and $\logit\pi=\log\frac\pi{1-\pi}$. Combining~\eqref{eq:indep} and~\eqref{eq:suff}, we find the \emph{normalization killer equation}:
\begin{align}
\label{eq:nke}
P(\tilde h\mid\tilde s,\pi) &= P(\tilde h \mid \tilde e, \tilde t, \tilde s,\Sset_{\tilde t},\Sset_{\tilde e}, \Sset_I, \pi,\Theta)
\end{align}
We have shown that if the scoring function properly computes the likelihood-ratio, using the \emph{same} model parameters that are assumed to have generated all the data, then everything but $\tilde s,\pi$ is irrelevant to inferring $\tilde h$. In this case, the raw score, $\tilde s$, is also the final score.

When i-vector PLDA is trained on large amounts of in-domain data \cite{garcia2014unsupervised,shum2014unsupervised}, it seems we are close to this ideal, because score normalization usually does not improve accuracy under those circumstances, although calibration may still be required.

\subsection{Non-ideal circumstances}
It is not hard to come up with excuses for score normalization. The above ideal circumstances are fragile. If, for any reason, the scoring function is different from~\eqref{eq:llr}, then the special property~\eqref{eq:suff} does not apply and dependency flows from $\tilde h$ through $\tilde e$ and $\tilde t$ to the cohort scores. There are several reasons why the scoring function can be different from~\eqref{eq:llr}, such as:
\begin{itemize}
	\item The recognition model is too complex and we have to resort to approximate scoring.
	\item The recognition model is a poor fit to the data.
	\item The recognition model has been trained on too little data and the parameter estimate is inaccurate.
	\item There is data-set shift between the training data and the trial-at-hand, where we may encounter, for example, new languages and recording channels.
\end{itemize}
In our experiments on RSR 2015 below, the first and possibly the second reasons explain the need for score normalization. All of these circumstances mean that, in practice, the true $\Theta$ that generated the data is not available for scoring and should therefore be considered hidden. A principled inference for $\tilde h$ must then be of the form:
\begin{align}
\label{eq:norm_post}
P(\tilde h \mid \tilde s,\Sset_{\tilde t},\Sset_{\tilde e}, \Sset_I, \pi)
\end{align}
The take-home message is simply that the cohort scores \emph{should} be used in some way.

\section{Probabilistic score normalization}
\label{sec:psn}
Once committed to score normalization, we need to find a way to compute $P(\tilde h \mid \tilde s,\Sset_{\tilde t},\Sset_{\tilde e}, \Sset_I, \pi)$. This generalizes score calibration~\cite{cal}, which merely computes $P(\tilde h\mid\tilde s,\pi)$. For reasons of simplicity and computational efficiency---just as in calibration---we will be ignoring the real, complex, mechanisms that produce the scores and instead resort to a simple score-space model.

A brute-force solution could discriminatively train a non-linear binary classifier with a cross-entropy criterion. This would be a non-linear logistic regression that processes all of the score inputs to produce a log-likelihood-ratio output score. The disadvantages include: (i) difficulty in designing the functional form, especially for cohorts of variable sizes, (ii) how to avoid specialization to the cohorts used in training, (iii) a large parameter count, which increases vulnerability to over-fitting and to data-set shift.

We choose the generative alternative, where we explicitly model dependencies between the scores and between scores and hypotheses. This has the advantage over the discriminative solution, that once the model has been defined, the functional form for the score can be derived by following the rules of probability theory. Variable sizes of cohorts present no problem. As we will show, we can define a model with very few trainable parameters. The danger of specializing on a given cohort choice at training time is not encountered because there is no notion of a cohort at training time.

One way to induce a general dependency between scores is to simply do Bayesian score calibration~\cite{bayescal}, where the score calibration parameters are marginalized out. This model, however, is too simple because it has no concept of \emph{trial sides} that induce structured dependencies between subsets of scores.

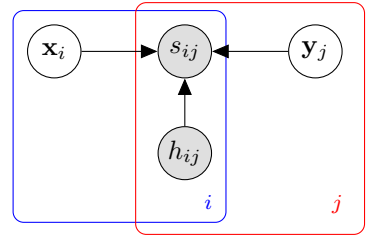
\begin{figure}[htb!]
  \centering
	\begin{tikzpicture}
	\node[obs] (s) {$s_{ij}$};
	\node[latent, left = of s] (x) {$\xvec_i$};
	\node[latent, right = of s] (y) {$\yvec_j$};
	\node[obs, below = 2 em of s] (h) {$h_{ij}$};
	\plate[draw=blue,inner sep = 5pt] {} {(x)(s)(h)} {\color{blue}$i$};
	\plate[draw=red,inner sep = 8pt] {} {(y)(s)(h)} {\color{red}$j$};
	\edge{x,y,h}{s}
	\end{tikzpicture}
  \caption{Normalization model at training time.}
  \label{fig:gm2}
\end{figure}

We decided, instead, to roughly try to emulate the look-and-feel of ZT-norm, with a model that associates a hidden variable, $\xvec_i$ with every enrollment representation, $e_i$, and another hidden variable, $\yvec_j$, with every test speech representation $t_j$. The hidden variables can be scalars, or smallish vectors. At training time this model can be represented as in figure~\ref{fig:gm2}.

The training data is a $K$-by-$L$ matrix of scores, $\{s_{ij}\}_{i,j=1}^{K,L}$, with associated hypothesis labels $h_{ij}$, obtained by scoring $K$ enrollments against every one of $L$ test representations. The model is defined via the joint distribution:
\begin{align}
\prod_{i=1}^K\prod_{j=1}^L P(\xvec_i)P(\yvec_j)P(s_{ij}\mid \xvec_i,\yvec_j,h_{ij})
\end{align}
In our experiments below we generalize training to make use of multiple independent score matrices. The training procedure depends on the model complexity. For the linear-Gaussian model presented below, we use an EM algorithm.

\subsection{Scoring}
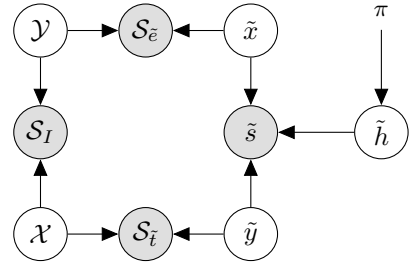
\begin{figure}[htb!]
  \centering
	\begin{tikzpicture}
	\node[const] (Theta) {};
	\node[obs, right = of Theta] (tildes) {$\tilde s$};
	\node[latent, above = 2 em of tildes] (tildee) {$\tilde x$};
	\node[latent, below = 2em of tildes] (tildet) {$\tilde y$};
	\node[obs, left = of Theta] (X) {$\Sset_I$};
	\node[latent, above = 2 em of X] (T) {$\Yset$};
	\node[latent, below = 2 em of X] (E) {$\Xset$};
	\node[obs] (St) at(tildet -| Theta) {$\Sset_{\tilde t}$};
	\node[obs] (Se) at(tildee -| Theta) {$\Sset_{\tilde e}$};
	\node[latent, right = of tildes] (tildeh) {$\tilde h$};
	\node[const, inner sep = 5pt, above = of tildeh] (pi)  {$\pi$};
	\edge{pi}{tildeh};
	\edge{tildeh}{tildes};
	\edge{tildee,tildet}{tildes};
	\edge{E,T}{X};
	\edge{T,tildee}{Se};
	\edge{E,tildet}{St};
	\end{tikzpicture}
  \caption{Normalization model at runtime.}
  \label{fig:gm3}
\end{figure}

At runtime, the model is rearranged as in figure~\ref{fig:gm3}, where the cohort score sets are $\Sset_{\tilde e},\Sset_{\tilde t},\Sset_I$, as defined above. The hidden variables are represented as $\Xset=\{\xvec_i\}$ and $\Yset=\{\yvec_j\}$. The observed data can still be viewed as a rectangular score matrix, where $\Sset_{\tilde e}$ and $\Sset_{\tilde t}$ occupy the same row and column, respectively, as $\tilde s$. We assume all hypothesis labels are given except for $\tilde h$, which must be inferred as:
\begin{align}
P(\tilde h =\texttt{target}\mid \tilde s,\Sset_{\tilde t},\Sset_{\tilde e}, \Sset_I, \pi)
&= \sigma(s^* + \logit \pi)
\end{align}
where $s^*$ is the \emph{normalized score}, computed as the log-likelihood-ratio:
\begin{align}
\begin{split}
s^* &= \log\frac{P(\tilde s,\Sset_{\tilde t},\Sset_{\tilde e}, \Sset_I\mid\tilde h =\texttt{target})}
{P(\tilde s,\Sset_{\tilde t},\Sset_{\tilde e}, \Sset_I\mid\tilde h =\texttt{non-target})} \\
&= \log\frac{P(\tilde s\mid\Sset_{\tilde t},\Sset_{\tilde e}, \Sset_I,\tilde h =\texttt{target})}
{P(\tilde s\mid\Sset_{\tilde t},\Sset_{\tilde e}, \Sset_I,\tilde h =\texttt{non-target})} \\
\end{split}
\label{eqn:normllr}
\end{align}
Evaluation of this expression requires marginalization over the hidden variables, the complexity of which depends on the model. For the linear-Gaussian model presented below, we find a closed-form solution.

\subsection{Linear-Gaussian normalization model}
\def\Hset{\mathcal{H}}
\newcommand{\embrace}[1]{\left\lbrace{#1}\right\rbrace}
\newcommand{\expect}[1]{\left\langle{#1}\right\rangle}

\newcommand{\ND}[3]{\mathcal{N}\left({#1}\mid {#2},{#3}\right)}

\newcommand{\dett}[1]{\left|{#1}\right|}
\newcommand{\tr}[1]{\textnormal{tr}\left({#1}\right)}
\def\muvec{\boldsymbol{\mu}}
\def\Lmat{\boldsymbol{\Lambda}}
\def\sij2{\sigma_{ij}^2}
\def\gammavec{\boldsymbol{\gamma}}
\def\alphavec{\boldsymbol{\alpha}}
\def\betavec{\boldsymbol{\beta}}
\def\Amat{\mathbf{A}}
\def\Bmat{\mathbf{B}}
\def\Cmat{\mathbf{C}}
\def\Imat{\mathbf{I}}

To enjoy closed-form training and scoring, we implemented a simple, linear-Gaussian~\cite{roweis1999} version of the general model described above. The hidden variables $\xvec_i$ and $\yvec_j$ are multivariate random variables of dimension $D$ and have independent standard Gaussian priors. In the likelihood, the score means are linear functions of the hidden variables:
\begin{align*}
P(s_{ij}\mid \xvec_i, \yvec_j, h_{ij}) &= \ND{s_{ij}}{ \mu_{ij} + \alphavec_{ij}^T \xvec_i + \betavec_{ij}^T \yvec_j }{\sigma_{ij}^2}
\end{align*}
where all of $\mu_{ij}\in\{\mu_{tar},\mu_{non}\}$, $\alphavec_{ij}\in\{\alphavec_{tar},\alphavec_{non}\}$, $\betavec_{ij}\in\{\betavec_{tar},\betavec_{non}\}$ and $\sigma_{ij}\in\{\sigma_{tar},\sigma_{non}\}$ agree with $h_{ij}\in\{\texttt{target},\texttt{non-target}\}$. The model therefore has a total of $4+4D$ trainable parameters. Notice if $\alphavec_{ij}=\betavec_{ij}=\boldsymbol{0}$, the model simplifies to a Gaussian calibration model~\cite{cal}.

In training, the hidden-variable posterior is essential to the EM algorithm. Due to \emph{explaining away}, the hidden variables are dependent in the posterior. If we denote by $\zvec$, the vector of stacked hidden variables, $\xvec_i$ followed by the $\yvec_j$, the posterior can be written as the multivariate Gaussian:
\begin{align}
  P(\Xset,\Yset \mid \Sset, \Hset) &= \ND{\zvec}{ \muvec_z }{\Lmat_z^{-1}}
\end{align}
where $\Sset$ and $\Hset$ denote all scores and labels and where
\begin{align*}
 \Lmat_z  &= \left[ \begin{array}{cc}
   \Amat & \Cmat \\
   \Cmat' & \Bmat
 \end{array}\right], &
\muvec_z  &= \Lmat_z^{-1} \left[ \begin{array}{c} \gammavec_x \\ \gammavec_y  \end{array}\right]
\end{align*}
where $\Amat,\Bmat$ are block-diagonal and where $\Amat,\Bmat,\Cmat, \gammavec_x,\gammavec_y$ have elements:
\begin{align*}
\Amat_{ii} &= \Imat + \sum_j\frac{ \alphavec_{ij}\alphavec_{ij}^T}{\sij2}, &
\Cmat_{ij} &=  \frac{\alphavec_{ij} \betavec_{ij}^{T}}{\sij2}, \\
\Bmat_{jj} &= \Imat + \sum_i\frac{ \betavec_{ij}\betavec_{ij}^T}{\sij2} \\
 \gammavec_{x,i} &= \sum_j \frac{s_{ij} - \mu_{ij}}{\sij2}\alphavec_{ij} , &
 \gammavec_{y,j} &= \sum_i \frac{s_{ij} - \mu_{ij}}{\sij2}\betavec_{ij}
\end{align*}
The model is trained using an EM algorithm with minimum divergence \cite{brummer2010:vbem}.

The runtime scoring formula can be obtained conveniently in terms of the hidden-variable posterior, by application of the \emph{candidate's formula} \cite{besag1989}. The numerator and denominator of \eqref{eqn:normllr} are the marginal distributions for all scores ($K\times L$ cohort scores + trial at hand $\tilde{s}$), $\Sset$, given all labels (cohort + hypothesized $\tilde h$), $\Hset$:
\begin{align*}
\begin{split}
  \log P(\Sset\mid\Hset) &= -\frac{1}{2} \sum_i^{K+1}\sum_j^{L+1} \left( \frac{ (s_{ij} - \mu_{ij})^2}{\sij2} + \log {2\pi\sij2} \right) \\
 & +\frac{1}{2}\muvec_z^T\Lmat_z \muvec_z - \frac{1}{2}\log\dett{\Lmat_z}. \\
\end{split}
\label{eqn:marginal}
\end{align*}
In situations where all scores between the cohort and the trial-at-hand are non-targets, we only need to pre-compute and Cholesky-factorize the posterior precision matrix twice, once for each value of $\tilde h$. This gives fast, vectorized scoring that can be applied to large sets of evaluation trials.

\section{Experiments}

We evaluate our score model on the RSR 2015 corpus \cite{larcher2015}.  Part I of the corpus consists of $143$ female and $157$ male speakers speaking 30 prompted phrases during 9 recording sessions. The dataset is split equally into background, development and evaluation sets with non-overlapping speakers. Our text-dependent speaker recogniser is a GMM-UBM system with eigenchannel compensation, similar to \cite{kenny2014}. Gender-dependent systems are trained on the background and development sets.

We train a score normalization model on the development set for each gender. Each model is trained on a collection of $30$ phrase-specific score matrices to exclude cross-phrase trials.  At runtime, the same development sets are used for gender-dependent score normalization cohorts.  We also report results only on the same-phrase trials. 

\begin{figure}[htb!]
  \centering
  \begin{tikzpicture}
    \begin{axis}[
       xtick       = {-4.26489,-3.29053,-2.57583,-2.05375,-1.64485,-1.28155,-0.841621,-0.253347},
       xticklabels = {0.001,0.05,0.5,2,5,10,20,40},
       xmin        = -4.5,
       xmax        = -0.85,
       ytick       = {-3.29053,-2.87816,-2.57583,-2.32635,-2.05375,-1.64485,-1.28155,-0.841621,-0.253347,0.841621},
       yticklabels = {0.05,0.2,0.5,1,2,5,10,20,40,80},
       ymin        = -3.3,
       ymax        = -0.8,
       xlabel      = False Alarm Probability (\%),
       ylabel      = Miss Probability (\%),
       legend style = {legend pos=north east,font=\tiny},
       legend cell align = left,
      ]
        \addplot[color=black,style=dashed,line width=0.5pt] table[x=pfa,y=pmiss]{male_det_data_raw.dat};
        \addplot[color=blue,style=dashed, line width=0.5pt] table[x=pfa,y=pmiss]{male_det_data_zt2.dat};
        \addplot[color=green,style=dashed,line width=0.5pt] table[x=pfa,y=pmiss]{male_det_data_new1d.dat};
        \addplot[color=red,style=dashed,  line width=0.5pt] table[x=pfa,y=pmiss]{male_det_data_new2d.dat};
        \addplot[color=black,line width=0.5pt] table[x=pfa,y=pmiss]{female_det_data_raw.dat};
        \addplot[color=blue, line width=0.5pt] table[x=pfa,y=pmiss]{female_det_data_zt2.dat};
        \addplot[color=green,line width=0.5pt] table[x=pfa,y=pmiss]{female_det_data_new1d.dat};
        \addplot[color=red,  line width=0.5pt] table[x=pfa,y=pmiss]{female_det_data_new2d.dat};
        \legend{{male, raw, EER=$1.24\%$}, {male, ZT, EER=$0.73\%$}, {male, LGSM 1D, EER=$0.8\%$}, {male, LGSM 2D, EER=$0.76\%$}, {fem, raw, EER=$0.53\%$}, {fem, ZT, EER=$0.27\%$}, {fem, LGSM 1D, EER=$0.38\%$}, {fem, LGSM 2D, EER=$0.27\%$}}
      \end{axis}
  \end{tikzpicture}
  \caption{Score normalization results the RSR 2015 corpus}
  \label{fig:dets_rsr}
\end{figure}
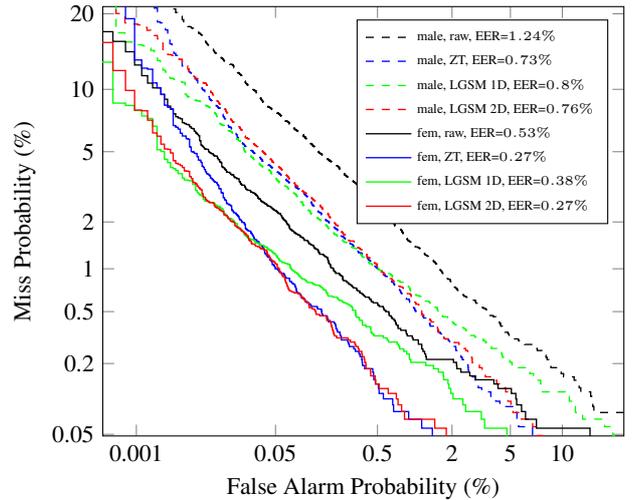

It is interesting to note that the hidden variables account for a significantly higher portion of the target score variance ($\nu_{tar} = \alphavec_{tar}^T\alphavec_{tar} + \betavec_{tar}^T\betavec_{tar}$) than for non-target scores:
\begin{center}
\begin{tabular}{l|cc|cc}
  & $\sigma_{non}^2$ & $\nu_{non}$ & $\sigma_{tar}^2$ & $\nu_{tar}$\\
\hline
Female & $23.4$   &  $3.5$ & $58.7$   & $55.0$ \\
Male   & $18.9$   &  $4.1$ & $55.7$   & $66.3$  \\
\end{tabular}
\end{center}
In Figure \ref{fig:dets_rsr} we compare raw, un-normalized scores and ZT-norm against our score normalization with scalar ($\texttt{LGSM 1D}$) and $2$-dimensional hidden variables ($\texttt{LGSM 2D}$).  Our score model improves results everywhere relative to raw scores and performs similar or better than ZT-norm in spite of the limitation that our model can only shift scores.

\subsection{Calibration}
Alert readers may wonder whether this model naturally gives calibrated scores. Since the normalized scores are log likelihood-ratios, this should theoretically be the case. As a sanity check, we found this to hold for synthetically generated data. On RSR data, our normalized scores were better calibrated than raw and ZT-norm scores in the low false-alarm region, but this was not consistent at all operating points. This suggests our model may not have enough capacity to properly model all aspects of real scores. We hope future work, with more sophisticated models,  may improve this.

\section{Conclusion}

We proposed a framework for reasoning about, and designing score normalization models in a principled, probabilistic manner. Our first implementation of a simple linear-Gaussian score model excercises this approach on real data and achieves results comparable to existing score normalization algorithms.

Future work includes developing non-linear score models in which hidden variables can have non-linear effect on the scores.

\section{Acknowledgements}
We thank the Brno University of Technology Speech Group for helpful discussions during our collaboration in NIST SRE'16.

\bibliographystyle{IEEEtran}

\bibliography{\jobname}

\end{document}